\RequirePackage{luatex85}
\documentclass[twocolumn]{ceurart}
\usepackage[utf8]{inputenc}
\usepackage{amssymb}
\usepackage{amsmath}
\usepackage{amsthm}
\usepackage{xspace}
\usepackage{graphicx} % for \rotatebox
\usepackage{listings} % for listing code
\usepackage{mathrsfs}
\DeclareMathAlphabet{\mathantt}{OT1}{antt}{li}{it}
\DeclareMathAlphabet{\mathpzc}{OT1}{pzc}{m}{it}
\usepackage[normalem]{ulem} % For striking text
\usepackage{xcolor}
\usepackage{proof}
\usepackage{url}
\usepackage{times}
\usepackage{soul}
\usepackage{url}
\usepackage{hyperref}
\usepackage{caption}
\usepackage{booktabs}
\usepackage{algorithm}
\usepackage{algorithmic}
\usepackage{cleveref}
\newcommand{\caisarurl}{\url{https://git.frama-c.com/pub/caisar}\xspace}

\newcommand{\mcl}[1]{} % multi line comment
\newcommand{\mlc}[1]{} % multi line comment

\newcommand{\zomit}[1]{}

\newcommand{\eg}{\emph{e.g.,\xspace}\xspace}

\newcommand{\InSpace}{\mathcal{X}} %
\newcommand{\VerifProb}{\mathcal{V}} %

\begin{document}
\copyrightyear{2022}
\copyrightclause{Copyright for this paper by its authors.
	Use permitted under Creative Commons License Attribution 4.0
	International (CC BY 4.0).}

\conference{The IJCAI-ECAI-22 Workshop on Artificial Intelligence Safety (AISafety 2022),
	July 24-25, 2022, Vienna, Austria}

\title{CAISAR: A platform for Characterizing Artificial Intelligence
	Safety and Robustness}

\author[1]{Julien Girard-Satabin}[
	orcid= 0000-0001-6374-3694 ,
	email=julien.girard2@cea.fr,
]
\fnmark[1]
\author[1]{Michele Alberti}[
	email=michele.alberti@cea.fr,
]
\fnmark[1]
\author[1]{François Bobot}[
	email=francois.bobot@cea.fr,
]
\author[1]{Zakaria Chihani}[
	email=zakaria.chihani@cea.fr,
]
\author[1]{Augustin Lemesle}[
	email=augustin.lemesle@cea.fr,
]
\address[1]{
	Université Paris-Saclay, CEA, List, F-91120, Palaiseau, France
}

\fntext[1]{These authors contributed equally.}


\maketitle

\begin{abstract}
	We present CAISAR,
	an open-source platform under active
	development for the
	characterization of AI systems' robustness and safety.
	CAISAR provides a unified entry point for defining verification
	problems by using WhyML, the mature and expressive language of the Why3
	verification platform.
	Moreover, CAISAR orchestrates and composes state-of-the-art
	machine learning verification tools which, individually,
	are not able to efficiently handle all problems but, collectively,
	can cover a growing number of properties. Our aim is to assist,
	on the one hand, the V\&V process by reducing the burden of choosing the
	methodology tailored to a given verification problem,
	and on the other hand the tools developers by factorizing useful features
	-- visualization, report generation, property description --
	in one platform.
	CAISAR will soon be available at \caisarurl{}.
\end{abstract}

\section{Introduction}
The integration of machine learning programs as
components of critical systems is said to be bound to
happen; initiatives from various private and govermental
actors
(\eg US' NSF funding for Trustworthy
AI\footnote{https://www.nsf.gov/pubs/2022/nsf22502/nsf22502.htm},
France's Grand Défi IA de confiance\footnote{https://www.gouvernement.fr/grand-defi-securiser-certifier-et-fiabiliser-les-systemes-fondes-sur-l-intelligence-artificielle}) are a consequence of that fact.
Trusting such programs is thus becoming a
crucial issue, both on technical and ethical
sides.

A possible approach to trust is formal test and verification, a broad set
of techniques and tools that have been applied to software safety for several
decades. These formal methods build on sound
mathematical foundations to assess the behaviour of programs
in a principled way, be it for generating tests or providing proven guarantees.
For the last few years, several independent works have started
to investigate the possible applications of formal verification
to machine learning program verification and its
limitations. This led to what could be characterized as a
Cambrian explosion of tools aiming to solve a particular
subset of the machine learning verification field.
In less than five
years, more than 20 tools were produced, upgraded or
abandoned. These tools use different techniques, different
input formats, handle different ML artefacts
and, most importantly, have varying performances depending on the problems.
Our goal of orchestrating multiple tools aims at maximizing the property
coverage in the context of a validation and verification process.

%Even if initiatives such as the Verification of Neural Network Competition (VNN-COMP) ~\cite{bak2021second} aim to provide a fair comparison between tools, there is still work to be done. For instance, beside of well-known benchmarks, using a specific tool still requires - most of the time - expressing a verification problem in a particular specification language. Even more, some tools may not be suitable at all for the verification problem of interest. 

To this end,
we present a platform dedicated to Characterizing Artificial
Intelligence Safety and Robustness (CAISAR) that aims to
unify several formal methods and tools, at the input,
through the use of a mature and expressive property description and proof
orchestration language, at the output,
through the factorization of features
such as visualization and report generation, and at the usage,
through shared heuristics and interconnections between tools.

The answer to the question ``To what extent can I trust my machine
learning program?'' has many components,
ranging from data analysis to decision explainability.
One such important components is dealing with verification and
validation, and we wish to make CAISAR an important element
in the safety toolbelt by covering these applications.

In the following, we will first present the design principles
of CAISAR and state its main goals. We will follow by a
description of its most prominent features, as well as its
limitations. We will then explain the position of CAISAR
regarding other tools for formal verification of machine
learning programs, and conclude by presenting some future
work and possible research problems.

\section{Core principles of CAISAR}

The aim for CAISAR is to provide a verification environment
for Artificial Intelligence (AI) based systems tailored to different needs.
The profusion of tools for AI-programs certification
offers numerous possibilities, from the choice of
technology (formal methods, test generation) to the scope
of properties to check (coverage, suitability to a given distribution,
robustness). However, with increased possibilites comes
the burden of choice.
Which method better suits a given use case? Are the results provided by this
particular method
trustworthy enough? How to bring trust in the process of
selecting, tailoring and computing results of a given tool?
How to evaluate a given tool against others? Those are the
questions that we aim to answer with CAISAR.

\subsection{Compatibility with existing and future methods}

The first principle of CAISAR is to ease this burden of
choice by automating parts of it. CAISAR aims to provide a
unique interface for vastly different tools, with a single
entry point to specify verification goals. Choosing which tool to
use is an informed decision that may not be relevant for
the user; the goal is to provide an actionable answer on
the safety of the system, by using whatever tool is
suitable for the problem. Ideally, the user should not
be bothered with deciding which tool is suitable for their use case:
CAISAR will
automatically figure out how to express the given property
to suit verifiers. As AI systems pipelines are becoming
more and more complex, it is crucial for CAISAR to handle
this complexity. Currently, CAISAR supports
neural networks and support vector machines,
and an industrial benchmark of an ensemble model (NN-SVM),
which we are unable to further discuss,
is being used as a concrete real-word use-case.

\subsection{Common modelling and answers}

%je ne sais pas si le fait de justifier cette sous-section par les bugs
% est la meilleure interface à utiliser pour justifier la plateforme

Existing verifiers rely on different decision procedures, \eg Mixed
Integer Linear Programming (MILP), abstract interpretation,
or Satisfaction Modulo Theory (SMT) calculus.
Modelling a verification problem using these frameworks
require different skills and is time-consuming; for
instance, some modelling choices made for MILP may not be applied
under SMT. Moreover, even if one succeeds in phrasing a
verification problem under multiple decision procedures,
the different results may not be immediately comparable.

CAISAR aims to provide a common ground for inputs and
outputs, which will lead to an easier comparison, lower
time consumption and an informed
decision. Furthermore, collecting and presenting the user
with multiple answers from different techniques can provide
additional confidence on the studied system. In order for users
to trust CAISAR as well, it needs to rely on well-known and
approved principles and technologies. It is developed in OCaml, a
strongly typed programming language, used to develop tools
for program verification and validation. Such tools include
Coq~\cite{huet2002coq}, a proof assistant that for instance
used to develop CompCert\cite{leroy2016compcert}, a C compiler that is
guaranteed to output C-ANSI compliant source code,
Frama-C~\cite{baudin2021dogged}, a platform for the static
analysis of C code, and Why3~\cite{filliatre2013why3}, a platform for
deductive verification of programs.

\subsection{Tools composition}

Some works are starting to combine multiple
techniques~\cite{singh2019abstract} for their analysis,
using an exact MILP solver to refine bounds
obtained by abstract interpretation.
Our goal with CAISAR is to bring tool composition to another
level. For instance, metamorphic transformations could generate
different input space partitions for formal verifiers. A
reachability analysis tool could be called numerous times
with tighter bounds until reaching a precise enough answer.
Coverage testing objectives could be extracted from
reachability analysis tools and fed to test generators.
CAISAR will be more than the sum of its part, allowing
communication between vastly different tools to provide
faster and more accurate answers.

\subsection{Automatic proposal of verification strategies}
A long-term goal for CAISAR is to provide a reasoning engine
where past verification problems processed by CAISAR can inform next
ones, gradually building a knowledge base that is suitable
for the specific needs of the user.
CAISAR will also implement
its own built-in heuristics to supplement specialized
programs that do not implement them.

\section{Architecture and features}

\begin{figure}[!h]
	\begin{center}
		\includegraphics[height=0.45\textheight]{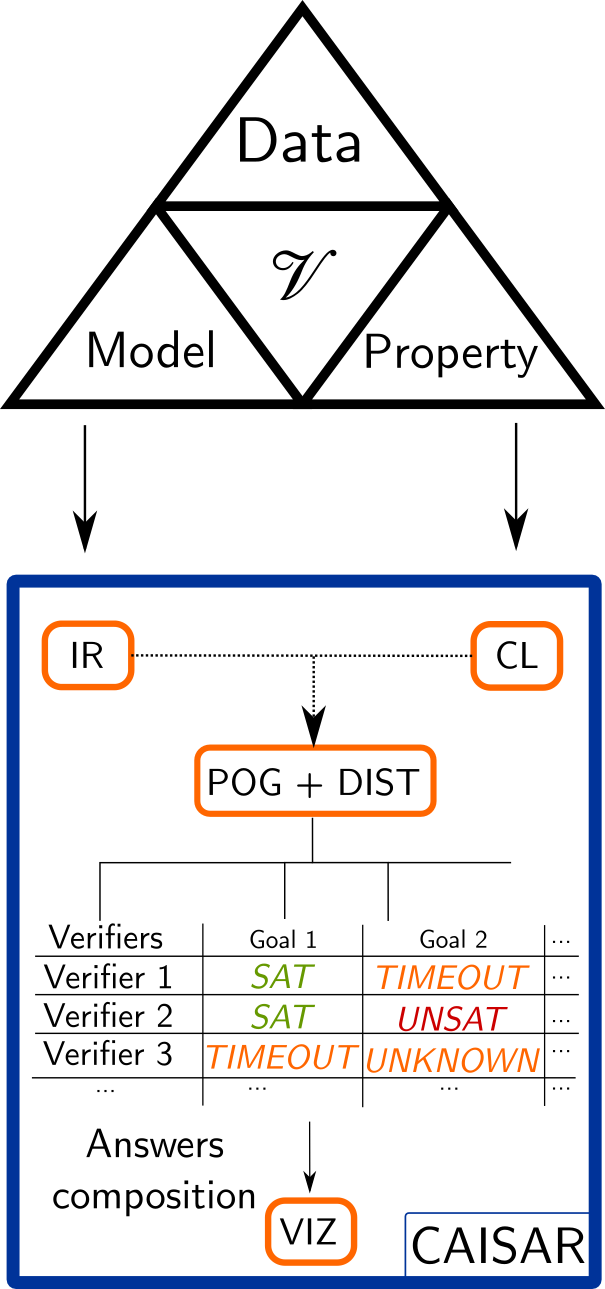}
	\end{center}
	\caption{CAISAR overall architecture}
	\label{fig:caisar}
\end{figure}

CAISAR's architecture
can be divided into the following functional blocks:
\begin{enumerate}
	\item A Common specification Language (CL)
	\item A Proof Obligation Generator (POG), associated with a
	      Dispatcher (DISP)
	\item An Intermediate Representation (IR)
	\item A visualization module (VIZ)
\end{enumerate}
See \cref{fig:caisar} for a visual depiction of dependencies
between blocks.

\subsection{Specification language and verification predicates}
A typical task for program verification involves to solve a
\emph{verification problem}. A verification problem consists
on checking that a program with a given set of inputs
is meeting certain expectations on its outputs.
More formally, let $\InSpace$ be an input space,
$\mathcal{P}(\cdot{})$ be a property on the input space,
$f$ be a program, and
$\mathcal{Q}(\cdot{}, \cdot{})$ be a property on the output space.
By property, we mean a statement that describes a desirable
behaviour for the program.
Let $\VerifProb = (\InSpace, f, \mathcal{P}, \mathcal{Q})$
be a \emph{verification problem}.
The goal is to verify the following property:
\[
	\forall x \in \InSpace, \mathcal{P}(x) \Rightarrow \mathcal{Q}(f(x), x)
\]
To write $\VerifProb${},
one needs to be able to express concisely and without ambiguity
each component: the program to verify
$f$,
the properties $\mathcal{P}$ and $\mathcal{Q}$, and the
dataset $\InSpace$.
To this end, CAISAR provides full support for the WhyML specification
and programming language~\cite{filliatre2013why3}. WhyML is a
language with static strong typing, pattern matching, types
invariants and inductive predicates. This gives WhyML programs a sound
semantic as logical propositions. WhyML is at the core
of the Why3 verification platform, and has been used
as an intermediate language for verification of Ada
programs~\cite{guitton2011why}.
This global expressiveness and safety allows to write
$\VerifProb${} once and for all, independently of the verifier.
For instance, \Cref{code:robust_predicate} shows the definition of
robustness against a perturbation of amplitude $\varepsilon$
using the $l_{\infty}$ distance within CAISAR's standard
library, and \cref{code:robustness_verification} the WhyML
file the user need to write in order to verify the
robustness of a given TestSVM against a perturbation of
amplitude $0.5$.
Note that all the necessary element to
define $\VerifProb{}$, namely $f$, $\InSpace$,
$\mathcal{P}$ and $\mathcal{Q}$, are defined in
those files: $f$ is the function \lstinline{TestSVM},
and
$\mathcal{Q}$ is the predicate itself.
Note that WhyML is not limited to the robustness
against a given perturbation property, often met in the
literature.
For instance, asserting that a neural network respect the
properties of being differentialy
private~\cite{abadi2016deep} or
respecting causal fairness~\cite{urban2019perfectly}
is something that
could be phrased as WhyML programs, since those properties
have a mathematical characterization.
Finally, WhyML does not constrain the form of
$\mathcal{P}$ nor $\mathcal{Q}$. In particular, it is
possible to define multiple verification goals in
the same $\VerifProb$, opening the way to subdivide
it into subproblems, and providing
answers at each step.

CAISAR then automatically translates $\VerifProb{}$
into a format supported by the selected verifiers, through
a succession of encoding transformations and simplifications.
For instance, some verifiers are best used when trying to
falsify the property: instead of checking
\[
	\forall x \in \mathcal{X}, \mathcal{P}(x) \Rightarrow
	\mathcal{Q}(f(x), x)
\]
they instead try to satisfy the negation
\[
	\exists x \in \mathcal{X}, \mathcal{P}(x) \land{}
	\lnot{} \mathcal{Q}(f(x), x)
\]
This transformation is embedded in CAISAR, when calling
Marabou: a verification
problem $\VerifProb$ can be transformed into an equivalent
one $\VerifProb{}^{'}$ that can be dispatched to Marabou.

\begin{figure}[bh!]
	\begin{center}
		\begin{lstlisting}[language=Caml]
type input_type = int -> t
type output_type = int
type model = {
          app :
          input_type -> output_type;
          num_input: int;
          num_classes: int
}

predicate dist_linf
  (a: input_type)
  (b: input_type)
  (eps:t)
  (n: int) = 
  forall i. 0 <= i < n ->
    .- eps .< a i .- b i .< eps

predicate robust_to
  (model: model)
  (a: input_type)
  (eps: t) =
  forall b. dist_linf a b eps
  model.num_input ->
  model.app a = model.app b
\end{lstlisting}
	\end{center}
	\caption{An example of a predicate in CAISAR's standard
		library: being ``robust to'' against a perturbation of
		amplitude $\varepsilon$. Here, the predicate defines
		$\mathcal{Q}(\mathcal{Y})$.}
	\label{code:robust_predicate}
\end{figure}

\begin{figure}[th!]
	\begin{center}
		\begin{lstlisting}[language=Caml]
  use TestSVM.SVMasArray
  use ieee_float.Float64
  use caisar.SVM

  goal G: forall a : input_type.
  robust_to svm_apply a (0.5:t)
\end{lstlisting}
	\end{center}
	\caption{Example verification problem specified to
		CAISAR. The program to verify is \lstinline{TestSVM},
		the input space is
		defined by the elements in \lstinline{a}, the output space
		is the result of the application of the function
		\lstinline{svm_apply}.}
	\label{code:robustness_verification}
\end{figure}

\subsection{Proof Obligations Generations for various tools}
CAISAR currently supports a variety of tools and techniques:
metamorphic testing, reachability analysis based on abstract
interpretation and constraint-based propagations. CAISAR can
analyze neural networks and support vector machines.
This versatility
allows for CAISAR to verify system components using different
machine learning architectures.

\subsubsection*{Marabou}
Marabou~\cite{katz2019marabou} is a deep neural network
verification complete verifier. Its core routine relies on a
modified simplex algorithm that lazily relaxes constraints on
piecewise linear activation functions. Marabou also makes use
of several heuristics that help speeding up the verification
procedure, like relying on tight convex overapproximations
~\cite{wu2022efficient} or sound overapproximations
~\cite{ostrovsky2022abstraction}. It can answer reachability
and conjunction of linear constraint queries.
Marabou ranked fifth at the VNN-COMP 2021. It is currently
in active development.

\subsubsection*{SAVer}
The Support Vector Machine reachability analysis
tool SAVer~\cite{ranzato2019robustness} is
specialized in the verification of support vector machines
(SVM), a popular machine learning algorithm used alongside
neural networks for classification or regression tasks.
SAVer can answer reachability queries, and supports a
variety of SVM configurations. This tool was selected for
support as, to the best of our knowledge, it is the first
one to deal with verification of SVM.

\subsubsection*{Alt-Ergo and SMTLIB compliant
	solvers}
Existing general purpose SMT solvers for program
verification like Alt-Ergo~\cite{conchon2018alt}
or Z3~\cite{demoura2008z3} all support
a standard input language, SMTLIB~\cite{barrett2016satisfiability}.
CAISAR leverages Why3 existing support for SMT solvers and
can translate neural network control flows directly into SMTLIB
compliant strings using its intermediate representation,
which allows the support of a variety of off-the-shelf
solvers.
Note that the VNNLIB standard, used in the VNN-COMP,
uses a subset of SMTLIB2, which paves the way for the
support of future tools in CAISAR.

\subsubsection*{Python Reachability Assessment Tool}
The Python Reachability Assessment Tool (PyRAT)
is a static analyzer targeting specifically neural networks.
It builds upon the framework of abstract
interpretation~\cite{cousot1977abstract} using abstract domains adapted for
the approximation of the reachable space in a neural network.
Three main domains are used: intervals with symbolic relations
as described in~\cite{li2019symbolic,wang2018efficient},
zonotopes~\cite{singh2018fast} and Deep poly
domain~\cite{singh2019abstract}.

For low dimensional inputs, PyRAT use input partitioning
as described in~\cite{wang2018efficient}, with heuristics tailored
to relational domains: the zonotope domain and the deeppoly domain
with backsubstitution.
Those heuristics allow the computation of a non-trivial
(\eg not just widest interval first) score ranking the inputs
by their estimated influence on the outputs.
PyRAT has comparable results to state-of-the-art analyzers on the widely
used ACAS-Xu~\cite{manfredi2016introduction} benchmark,
and outperforms the similar domains of ERAN on S-shape activations
functions such as the sigmoid or hyperbolic tangent
functions with specific approximations.

\subsubsection*{AIMOS: a Metamorphic testing utility}
AI Metamorphism Observing Software (AIMOS) is a software
developped at the same time as CAISAR,
aiming to provide metamorphic properties testing or perturbations
on a dataset for a given AI model.
Metamorphic testing is a testing technique relying on
properties symmetries and invariance on the operating
domain. See~\cite{chen2018metamorphica} for a comprehensive
survey on this approach.
AIMOS offers tools to derive properties from a set of transformations
on the inputs: given $\mathcal{P}$, $\mathcal{Q}$, $\mathcal{X}$ and a
transformation function $t_{\theta}: x \in \mathcal{X}
	\mapsto \mathcal{X}$, it generates a set of new properies
$\mathcal{Q}^{'}$ that are coherent with the transformation.
As an example, a symmetry on the inputs of a classification model could result on a
symmetry on the outputs; AIMOS would then automatically
modify the property to check against the symmetrical
labels.

AIMOS can generate test cases scenarios from the most
common input transformations (geometrical rotations, noise
addition); others can be added if necessary.
AIMOS was evaluated on a metamorphic property on the ACAS-Xu
benchmark. The aim of the property was to evaluate the
ability of neural networks trained on ACAS to generalize
with symmetric inputs.
Given a
symmetry on inputs, AIMOS generates the expected symmetrical
output, and tests models against the base and symmetrical
outputs.
See~\cref{tab:second-res} for results. AIMOS was able to
show that neural networks trained on ACAS have a low, but
noticeable sensitivity to symmetry on one input.
\begin{table}[!h]
	\caption{Average number of same answer
		for all 45 models of the ACAS-Xu benchmark, computed by
		AIMOS.
		First column denotes values presented in the
		benchmark. Second column denotes the percentage of
		identical answer given.}\label{tab:second-res}
	\begin{tabular}{cc}\hline
		Previous answer & Identical answer \\\hline
		COC             & 89.7\%           \\
		WL              & 95.9\%           \\
		WR              & 99.6\%           \\
		L               & 95.3\%           \\
		R               & 99.8\%           \\
		\bottomrule
	\end{tabular}
\end{table}

\subsection{Supported formats}
CAISAR supports all input formats used by its integrated
verifiers. Most verifiers require
either a framework-specific binary (Pytorch's pth,
Tensorflow tf), a custom
description language (NNet), or an Open Neural Network
eXchange (ONNX)~\footnote{https://onnx.ai/} file. CAISAR is able to parse any
of these input formats and extract useful metadata for the
building of the verification strategy. It can also output a
verification problem into
the SMTLIB~\cite{barrett2016satisfiability} format,
supported by all general purpose solvers, as well as in
the ONNX format. The VNN-Lib
initiative~\footnote{http://www.vnnlib.org/}
provides a standard format for verification problems that
relies on SMTLIB; thus CAISAR also supports VNN-Lib.
CAISAR aims for maximum interoperability,
and can be used as a hub to write and convert verification queries
adapted to different verifiers.
Additionally, verifiers sometimes require datasets to
verify properties against, especially reachability analysis
tools. As such, CAISAR currently supports
datasets as flattened features under a csv file, and RGB
images.

\subsection{Answer composition}
CAISAR currently offer two ways to compose verifiers.
First, CAISAR can launch several solvers on the same task and
compose their answer: it can then provide a summary stating which solver
succeeded and which one failed.
Second, CAISAR has the ability to verify pipelines that are
composed of several machine learning programs: for instance,
a pipeline composed of several neural networks, or a neural
network which outputs are processed by a SVM. CAISAR can be
used to state an overall verification goal, and to model
that the outputs of a block of the pipeline are the
inputs of another block.
More advanced methods of composition, such as automatic
subgoals generation or refinement by multiple analysis
constitute a promising research venue.

\section{Background \& related works}

In less than five years, a profusion of tools and techniques
leveraging formal verification to provide trust
on neural network
sprouted~\cite{katz2017reluplex,katz2019marabou,wang2018efficient,wang2021beta,singh2018fast,singh2019abstract,shi2020robustness,bak2021nnenum,henriksen2020efficient,dutta2017output,palma2021improved,urban2019perfectly,ehlers2017formal}.
See~\cite{urban2021review,liu2019algorithms} for more comprehensive surveys on
the verification and validation of machine learning
programs.

As for general purpose verification platforms, examples
include the Why3 deductive verification platform and the
Frama-C~\cite{baudin2021dogged} C static analysis platform.
We leverage multiple existing features of Why3,
such as the WhyML language support, transformation and rewriting
engine. Why3 and Frama-C both lack the interfaces and
tooling to handle neural network. Experiments we conducted involving the
EVA Frama-C plugin applied on simple reachability analysis
properties showed a lack of scalability that a naive python
reachability analysis tool, specialized in neural networks,
was able to overcome quickly. Conversion of neural networks
in C programs that were scalable for EVA presented difficult
challenges. The differing structure between C
programs and neural networks implies differing
verification problems. Thus, it seems more fruitful to investigate a
specialized platform for machine learning programs.

The ProDeep platform~\cite{li2020prodeep} aims to regroup
several verifiers under a single user interface. It provides a
single entry point, supports input formats and offers
numerous visualization tools. It does not aim to provide
other properties than those that are natively supported by
its embedded verifiers. It also supports a fixed set of
datasets. They make use of
DeepG~\cite{balunovic2019certifying} to generate constraints
for verifiers, effectively combining tools.
Their scope seems limited to neural
networks, whereas CAISAR currently supports neural networks
and support vector machines, and aims to support a wider set of
machine learning models.

The most similar work to CAISAR is the DNNV platform~\cite{shriver2021dnnv}.
As CAISAR, DNNV provides support to various state-of-the-art
verifiers. It similarly aims to be a hub for neural
network verification by supporting a wide range of input and
output formats, and by providing a modelling language for
properties specification and discharge to capable provers.
Their Domain Specific Language, DNNP, is built on Python;
while CAISAR's specification language, WhyML, is already
used in several formal verification platforms and provide
additional theoretical guarantees, which is a key component
to provide trust. As stated before, WhyML allows
specifying multiple verification goals in the same
verification problem, which helps modelling more complex use
cases.

The main difference between CAISAR and DNNV is that the
latter does not combine verifiers answers, that is to say
there is (at the time of writing) no feature that aims to
interoperate verifiers: from the DNNV
documentation\footnote{https://docs.dnnv.org/en/stable/}:
"DNNV standardizes the network and property input formats
to enable multiple verification tools to run on a single
network and property.
This facilitates both verifier comparison, and artifact re-use."
As verifiers are becoming more and
more sophisticated and specialized, combination of methods
will become even more fruitful, and we expect this to be a
key difference with DNNV.

\section{Conclusion \& future works}
As the field of machine learning verification is blooming,
choosing the right tool for the right verification problem
becomes more and more tedious. We presented CAISAR, a
platform aimed to alleviate this difficulty by presenting a
single, extensible entry point to machine learning
verification problem modelling and solving. Plenty of work still
needs to be done, however.

Altough CAISAR already integrates some state-of-the-art
tools, other verifiers that ranked high in the VNN-COMP are
on the way of integration. Such verifiers include
$\alpha,\beta$-CROWN~\cite{wang2021beta,xu2021fast}, who scored first
on said competition.

Another research venue would be the integration of neural network reparation
techniques such as~\cite{goldberger2020minimal}.
Corrective techniques would contribute to provide a feedback loop
composed of problem
specification, verification, fault identification
and correction proposal.

Various problem splitting heuristics based, for instance, on
~\cite{girard-satabin2021disco,bunel2020branch}
could be integrated into CAISAR to leverage
parallelism for verifiers that do not support them

Data is the cornerstone of modern machine learning systems,
and it is necessary to give tools to handle its complexity.
Support for more various data kinds, such as time series, is
a first step towards this direction. Integration of tools
for analyzing data \emph{in relation with} a program,
for instance out-of-distribution detection, is another
future work.

Finally, to further help the user to select the optimal set
of tools for its verification problem, a long-term goal of
CAISAR is to provide
a verification helper to design optimal queries
for verification problems based on previous runs.

\bibliography{laiser}

\end{document}